# Nature-Inspired Optimization Algorithms: Research Direction and Survey

ROHIT KUMAR SACHAN, IIT Kanpur, India

DHARMENDER SINGH KUSHWAHA, MNNIT Allahabad, India

Nature-inspired algorithms are commonly used for solving the various optimization problems. In past few decades, various researchers have proposed a large number of nature-inspired algorithms. Some of these algorithms have proved to be very efficient as compared to other classical optimization methods. A young researcher attempting to undertake or solve a problem using nature-inspired algorithms is bogged down by a plethora of proposals that exist today. Not every algorithm is suited for all kinds of problem. Some score over others. In this paper, an attempt has been made to summarize various leading research proposals that shall pave way for any new entrant to easily understand the journey so far. Here, we classify the nature-inspired algorithms as natural evolution based, swarm intelligence based, biological based, science based and others. In this survey, widely acknowledged nature-inspired algorithms namely- ACO, ABC, EAM, FA, FPA, GA, GSA, JAYA, PSO, SFLA, TLBO and WCA, have been studied. The purpose of this review is to present an exhaustive analysis of various nature-inspired algorithms based on its source of inspiration, basic operators, control parameters, features, variants and area of application where these algorithms have been successfully applied. It shall also assist in identifying and short listing the methodologies that are best suited for the problem.

## 1   INTRODUCTION

Recent past has witnessed a wide adoption of nature-inspired algorithms for diverse real-world optimization problems that include engineering experiments, scientific experiments and business decision making. These algorithms are based on randomization concept and draw inspiration from natural phenomenon. A few of the several nature inspired algorithms proposed till now, have proved to be very efficient. Many algorithms give adequate results, but no algorithm gives an admirable performance in solving of all the optimization problems. In other words, an algorithm may show good performance for some problems while it may perform poorly for other problems [219]. However, as compared to classical optimization techniques, nature-inspired algorithms obtain optimal solutions for a wider range of problem domains in a reasonably practical time.

In real world, optimization problem is categorized into two categories: single objective and multi-objective. In single objective problem, only one objective is optimized while, multi-objective problem focuses on more than one objective. Consequently, we have two types of optimization algorithms namely single objective optimization algorithms and multi-objective optimization algorithms. The term objective refers to an objective function which is a mathematical formulation of the optimization criteria of a problem. The objective function produces a numerical result based on input variables. The result of the objective function is known as fitness (or cost) and the number of variables given as input to the objective function is referred to as the dimension of objective function.

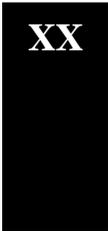

The main characterizing features of a good nature-inspired algorithm are a higher convergence rate, less processing time, an unbiased exploration and exploitation and less number of algorithm-specific control parameters. Convergence rate is the speed in terms of the number of iterations beyond which a repeated sequence is produced by an algorithm. This repeated sequence is known as a convergent sequence and is closer to the desired solution. The time required for execution of an algorithm is known as processing time. Exploration and exploitation are the elementary parameters of any nature-inspired algorithm. In exploration, exclusive new region of a search space is visited. In exploitation, only the neighborhood region of previously visited points of a search space is visited [36]. A good algorithm requires equilibrium between exploration and exploitation.

Generally, all nature-inspired algorithms require two types of controlling parameters: common control parameters (or regular parameters) and algorithm-specific control parameters (or dependent parameters) [159]. Common control parameters are the problem independent parameters like- population size, number of dimension, number of iterations, etc. On the other hand, algorithm-specific control parameters are the problem dependent parameters. In other words, the value of these parameters may differ from problem to problem like- GA requires a mutation and a crossover probability; PSO requires inertia weight and learning factors, etc. These dependent parameters may influence the performance of the algorithm. So a good nature-inspired algorithm should use minimum number of the algorithm-specific control parameters.

This paper presents an extensive overview and exhaustive analysis of various nature-inspired algorithms. The organization of the paper is as follows: Section 2 highlights the past related work. Classification of the nature-inspired algorithms is discussed in Section 3. Section 4 provides a broad review of various nature-inspired algorithms with their variants and applications. Section 5 enlists the comparative study of discussed algorithms based on its source of inspiration, basic operators, control parameters and features in chronological order. Section 6 outlines general conclusion.

## 2  RELATED WORK

Recently, many researchers have made an attempt to compare various existing evolutionary and nature-inspired algorithms. In paper [55], Elbeltagi et al. presented a comparative study of five evolutionary algorithms for continuous and discrete optimization. These algorithms are ant-colony, genetic, memetic, particle swarm and shuffled frog leaping. In another work, Parpinelli et al. [147] have reviewed the recently proposed swarm intelligence algorithms which also include a comparative analysis of ten algorithms based on source of inspiration, exploitation, exploration and communication model. Newly introduced algorithms like bat, cuckoo search and firefly are discussed by Sureja [206] along with the comparative result analysis of bat, cuckoo search, firefly, genetic and particle swarm for ten continuous and discrete optimization problems. In another paper [27], Binitha et al. presented a detailed literature survey and a comparision of various bio-inspired algorithms based on their representation, operators, control parameters and area of application. In [223], Yang discussed various search strategies and new challenges of nature-inspired meta-heuristic algorithms. In a related work Agarwal et al. [8] carried out a comprehensive review of twelve nature-inspired algorithms based on input parameters, evolutionary mechanism and applied application area while Kaur et al. [102] presented a comparative study of bat, cuckoo search, firefly and krill herd on the basis of their corresponding behaviour, objective function, features and area of application. A detailed insight of ant colony, artificial bee, evolutionary strategies, particle swarm, genetic algorithms and genetic programming has been outlined in the work of Dixit et al. [45].

## 3  CLASSIFICATION OF NATURE-INSPIRED ALGORITHMS

We categorize various Nature-Inspired Algorithms (NIAs) into five major categories based on the source of inspiration: natural evolution based, swarm intelligence based, biological based, science based and others. The natural evolution based algorithms are based on the basic principles of the theory of natural evolution. This theory is  known as "Darwinism". Swarm intelligence based algorithms are inspired by the collective behaviour of creatures like ants, bats, bees, cuckoos and fireflies. The biological based algorithms are motivated by the social behavioral pattern of biological systems. The science based algorithms are based on the scientific concepts. The algorithms that are inspired by any other natural phenomena fall into the category of others. Fig. 1 shows the classification of NIAs.

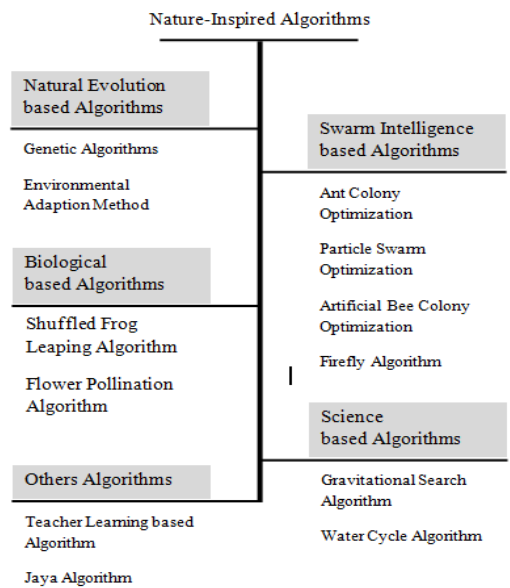

Fig. 1. Classification of nature-inspired algorithms

## 4  REVIEW OF NATURE-INSPIRED ALGORITHMS

The most prominent NIAs are Genetic Algorithms (GA), Ant Colony Optimization (ACO), Particle Swarm Optimization (PSO), Shuffled Frog Leaping Algorithm (SFLA), Artificial Bee Colony (ABC), Firefly Algorithm (FA), Gravitational  Search Algorithm (GSA), Cuckoo Search (CS), Bat Algorithm (BA), Environmental Adaption Method (EAM), Teacher Learning based Algorithm (TLBO), Flower Pollination Algorithm (FPA), Water Cycle Algorithm (WCA) and Jaya Algorithm. These algorithms are used for finding the optimal solution. These algorithms start with the search of optimal solution within search space of a randomly initialized population [147]. In each iteration, current population is replaced by a newly generated population. Fig. 2 shows the evolution timeline of various nature-inspired algorithms. Some of the widely recognized NIAs are examined in the subsequent sections.

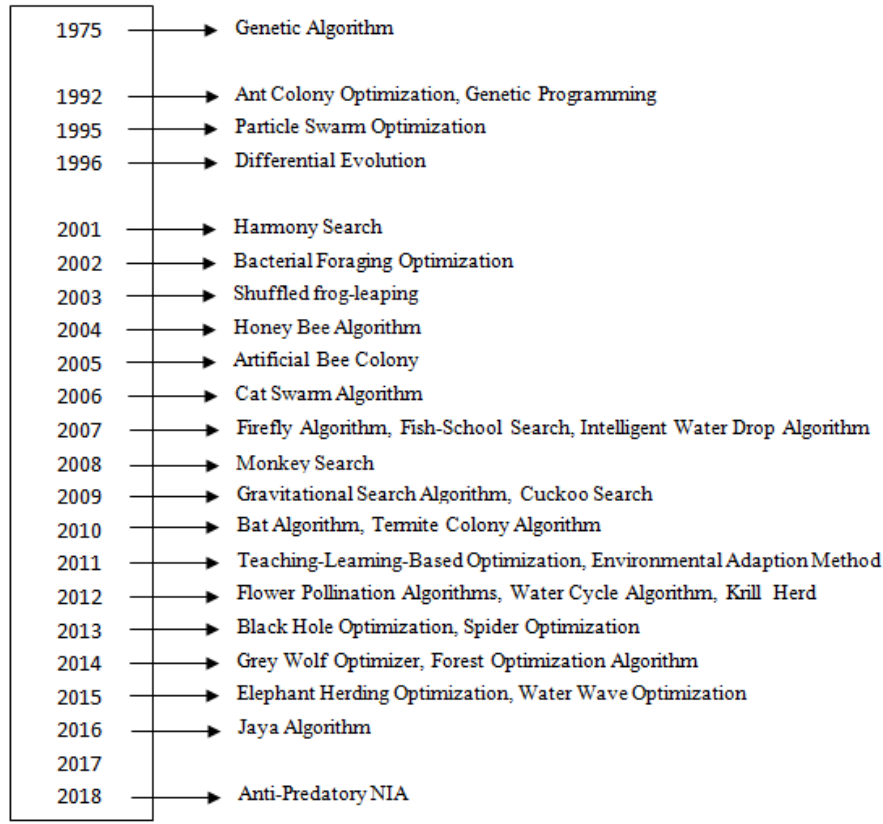

Fig. 2. Evolution timeline of nature-inspired algorithms

## 4.1  Natural Evolution based Algorithms

Natural evolution is the one of the oldest and well-known concept in the field of nature-inspired algorithms. These algorithms are inspired by theory of natural evolution. In this, only fittest individuals of current population are selected for next generation. The well-known natural evolution based algorithms are elaborated in the following sections.

### 4.1.1 Genetic Algorithms (GA)

Holland et al. [80] proposed the genetic algorithms at the University of Michigan. The concept of GA is based on the Darwin's theory of biological evolution- "Survival of the fittest". It states that only fittest individuals shall survive during the next generation while the unfit individuals shall be eliminated.

   Biological evolution is defined as a genetic change in population over time (or generations). Many times, these changes are very small and not noticeable. These changes occur at the gene level. The changes, which are fit for survival of genes, are only passed onto the next generation. During the biological evolution, various activities take place like- crossover and mutation of genes, selection of the fittest genes for the next generation.

In simulation of GA, individual genes (or solutions) are expressed in string format, called "chromosome". It uses three basic operators: crossover, mutation and selection [133]. In the evolution process, current population is replaced by new population, which has better average fitness than the previous generation. So the mean value of fitness of the next generation becomes fitter than its predecessor generation.

Various researchers have proposed different types of crossover, mutation and selection operators [193]. The principle behind the crossover and mutation is the same so as to modify or update the old chromosome and produce a new chromosome (or offspring). The main difference between crossover and mutation is that crossover operator is performed over two or more than two chromosomes, while the mutation operator is performed on a single chromosome. The fittest individuals are selected through selection operator.

The main parameters of GA are population size, number of generations, crossover probability, mutation probability and selection operator. The population size and number of generations are the common control parameters. The crossover probability, mutation probability and selection operator are the algorithm-specific control parameters. Length of chromosome and chromosome encoding method are also considered as algorithm-specific parameters. Due to crossover and mutation, GA has the ability of exploration and exploitation simultaneously.

The basic GA is sufficiently efficient. However variants of GA have been proposed to improve its effectiveness, efficiency and robustness. The comparative study of various existing variants is shown in Table 1. Numerous optimization problems have been successfully solved by GA. The recent applications are: software effort estimation [197], carpool service problem in cloud computing [84], VM placement in data centers [207], image enhancement and segmentation [151], water distribution system design [26], production ordering problem in an assembly process [188], medical image protection [145], wireless networks [142], vendor-managed inventory routing problem [146], parameter selection of photovoltaic panel [22] and QoS-aware service selection [44].

Table 1. Comparative study of GA variants

| Reference | Algorithm | Algorithm/s compared with | Application |
|---|---|---|---|
| [40] | Non-dominated Sorting GA II (NSGA II) | Pareto-Archived ES (PAES) and Strength-Pareto EA (SPEA) | Multi-objective test problems |
| [152] | Adaptive GA (AGA) | Project Scheduling Problem Library (PSPLIB) | Resource Levelling Problem (RLP) in project management |
| [121] | Multi-objective GA (MOGA) | Maximum applications scheduling algorithm and random scheduling algorithm | Scheduling in cloud computing |
| [124] | Real-coded GA | Nelder–Mead simplex, PSO and Bacterial Foraging (BF) | Brain images segmentation |
| [65] | Binary-Real Coded GA | Lagrangian Relaxation GA (GA-LR), Integer-Coded GA (ICGA), Matrix Real-Coded GA (MRCGA), Enhanced Simulated Annealing (ESA), SFLA | Unit Commitment (UC) problem |
| [236] | Hybrid GA/PSO | Existing IPPS methods | Process planning and scheduling |
| [178] | Simplified GA | COCOMO model | Software effort estimation |
| [13] | Elitism GA (EGA) | GA | Extreme Learning |

| | | | Machine (ELM) |
|---|---|---|---|
| [143] | GA with a decomposition strategy | GA | Unequal-Area Facility-Layout Problem (UA-FLP) |
| [120] | Hybrid Cuckoo Search-GA (CSGA) | ACO, PSO, Immune Based Algorithm (IBA) and Cuckoo Search (CS) | Benchmark problems and hole-making sequence optimization |

### 4.1.2 Environmental Adaption Method (EAM)

Environmental adaption method was proposed by Mishra et al. [129] for solving different benchmark functions. EAM is based on an improved theory of Darwinism. This theory does not consider the impact of current environmental conditions on the individuals. Due to this, the fitness improvement is very slow. In improved theory, impact of environmental conditions is considered for improving the fitness of individuals. In EAM, an individual adapts to environmental conditions to survive in a changing environment. Various observations state that during the evolution, average fitness of current generation always improves from the average fitness of previous generation. EAM improves the requirement of time and storage in comparison of GA and PSO [129].

The EAM simulates the improved Darwin's theory. For simulation, it uses three basic operators: adaption, mutation (or alteration) and selection [129]. Adaption operator is used to update the genome structure of individual. This modification depends on current fitness of individual and environmental fitness. For that, an exponential operator is used for damping the effect of change in the solution. As the solution moves towards the optimal solution, the ratio of fitness tends towards 1 as successive generation passes on. The mutation operator is applied on the individual to add the effect of environmental noise. The mutation operator just inverts the bits of individual genes with mutation probability. After this, best individuals are selected through a selection operator. During adaption, the new solution is calculated by equation 1.

$$new\_sol = \left( \alpha \times current\_sol^{fitness\,(current\_sol)/avg\_fitness} + \beta \right)/2^L - 1 \qquad (1)$$

where $\alpha$ and $\beta$ are random numbers and $L$ is the number of bits representing an individual.

The EAM requires three main control parameters. These are population size, number of generations and mutation probability. The population size and number of generations are the common control parameters. Mutation probability and size of an individual are the algorithm-specific control parameter. Due to adaption operator, fitness of individuals improves in a short duration of time. EAM has an unbiased exploration and exploitation [130]. The adaption operator exploits the neighborhoods of current solutions and simultaneously mutation operator explores the new solutions.

Some modified and updated versions of EAM are suggested by researchers. A comparative study of various existing variants of EAM is shown in Table 2. Limited number of applications of EAM have been found till date like- test case generation [130].

Table 2. Comparative study of EAM variants

| Reference | Algorithm | Algorithm/s compared with | Application |
|---|---|---|---|
| [129] | EAM | GA and PSO | Rastrigin and Schwefel function |
| [131] | Modified EAM (MEAM) | EAM and ePSO | Rosenbrock, Rastrigin, Griewank and Schwefel function |

| [130] | Improved EAM (IEAM) | GA and EAM | Several benchmark functions and Test case generation |
| [140] | Non-dominated Sorting EAM (NS-EAM) | NSGA-II | Two multi-objective function (Vanneta and Schaffer) |
| [210] | EAM for Dynamic environment (EAMD) | EAM | BBOB-2009 benchmark functions |
| [211] | Hybrid GA-EAM | PSO Time Variant Acceleration Coefficient (PSO-TVAC), Self-Adaptive DE (SADE) and EAM | Rosenbrock and Rastrigin function |
| [212] | EAM with Real parameter encoding for Dynamic environment (EAMD-R) | Several state-of-the-art algorithms | BBOB-2009 benchmark functions |

## 4.2 Swarm Intelligence based Algorithms

Swarm intelligence is a novel idea to inspire researchers to address optimization problem efficiently and effectively. The
se algorithms are based by the intelligent behaviour of the creatures like- ants, bees, birds, fishes, fireflies, bats and cuckoos. A large number of algorithms come under this category. The four widely used swarm intelligence based algorithms are described next.

### 4.2.1 Ant Colony Optimization (ACO)

ACO is an extended form of traditional construction heuristic [48]. The construction algorithms solve the problem in an incremental way. These start with an initial solution and iteratively a solution component is added randomly or greedily without backtracking. Greedy approach gives a better solution than the random approach, but it generates a limited number of solutions. The greediness is based on the profitability of the solution. The construction algorithms are the fastest approximation method, but often generate solutions that may be not optimal or high quality. With the help of the local search algorithms, solution can be improved. The local search algorithm explores the neighbors of the current solution and improves it. It finds the better neighbor solution if it exists.

In real life, the ants follow the stochastic construction approach and pheromone model for the search of the food. In this, new solution is generated by adding stochastic solution component in a particular solution. Stochastic component is a small random value which increases the randomness in search of optimal solution. Due to stochastic component, real ants discover a large number of solutions [48].

Dorigo [46] has proposed a new algorithm based on the ant's cooperative behavior. This is known as ACO. The ants move randomly in search of food. On discovering the food source, ants leave pheromone trails on the way back to their colony. This is used as a communication channel by the other ants. During the search of food, if any ant finds the pheromone path, they stop wandering and start to follow the pheromone path. The pheromone path signifies the presence of a food source. Multiple ants follow the same path and leave down pheromone trails, thus increasing the pheromone strength on the path. After a period of time, evaporation of the pheromone trail starts. The evaporation reduces the attractive strength of the deposited pheromone. In a comparison between the shorter path and longer path, a longer path has lesser pheromone density because it has more time for pheromone evaporation than the shorter path implying that shorter paths have distinctly higher pheromone density [47].

The basic ant colony metaheuristic follows the three steps [46, 47].

1. In first step, initial solutions are constructed by all ants.
2. In second step, solutions are improved by the local search algorithm.
3. In third step, pheromones are updated.

In simulation of ACO, ants memorize the traversed path and deposited pheromone. Ants try to search a path with minimum distance between colonies and the food source through random movements. The movement depends on the probability of solution components. The probability depends on the pheromone values and heuristic information, which are associated with each path. The probability of searching shortest path is inversely proportional to the route's distance. For each successive movement, the ants always choose the edge which has a higher probability of solution component. When all ants have completed their search, the pheromone amount of complete path is updated. The purpose of this is to improve the pheromone value of good solution and to reduce the pheromone value of bad solution. The worth of solution depends on the amount of pheromone levels on path associated with solutions. Different ACO variants may have different way to update pheromone.

The parameters required in ACO are number of ants, number of iterations, pheromone evaporation rate and amount of reinforcement. The number of ants and number of iterations are the common control parameters. The pheromone evaporation rate, heuristic information and amount of reinforcement are algorithm-specific control parameters. In ACO, a balanced exploration and exploitation can be achieved through the management of pheromone trails [49]. ACO is good for finding approximate solutions of hard optimization problems through graph or tree.

Dorigo et al. present recent advancement and application of ACO in paper [48] and recent advances in a successive paper [49]. Many customized and modified variants of ACO were proposed by various researchers. The comparative study of various existing variants of ACO is shown in Table 3. The recent applications of ACO are: software effort estimation [205], routing for mobile Ad-hoc network [31], spatial clustering algorithm [1], web service compositions in cloud computing [237], train routing selection problem [187], requirement selection in software development [41], energy-efficient networks [39], biometrics fusion [107], unsupervised probabilistic feature selection in pattern recognition [37] and blood vessel segmentation in retina diagnosis system [15].

Table 3. Comparative study of ACO variants

| Reference | Algorithm | Algorithm/s compared with | Application |
|---|---|---|---|
| [17] | Elitist Continuous ACO (ECACO) | ACO | Coastal aquifer management problem |
| [106] | Binary Ant System (BAS) | Continuous ACO (CACO), Continuous Interacting Ant Colony (CIAC), ACO and API algorithm | Unconstrained optimization problems |
| [235] | Continuous ACO (CnACO ) | Benchmark functions | Structural Health Monitoring (SHM) |
| [221] | Multi-Objective ACO (MOACO) | Ranked Positional Weight Method (RPWM), GA Artificial Immune Algorithm (AIS) | Mixed-model Assembly Line Balancing Problem (MALBP) |
| [116] | Hybrid GA-ACO | GA, ClustalW, Central-star algorithm and Horng's GA | Multiple Sequence Alignment (MSA) problem |

| [149] | Hybrid ACO-PSO (HACOPSO) | Tree-based PSO (PSOTREE) and Tree Growth based ACO Algorithm (TGBACA ) | QoS constrained multicast routing problem |
|---|---|---|---|
| [201] | Hybrid ACO-ANNs | ACO | Feature subset selection problem in field of medical diagnosis |
| [208] | Parallel ACO | ACO and Simulated Annealing (SA) | Resource job scheduling problem |

### 4.2.2 Particle Swarm Optimization (PSO)

In particle swarm optimization, the word swarm signifies the bird's flock or fish's school and the word particle denotes a bird in a flock or a fish in the school. PSO was based on the social behaviour of particles in swarms. This includes synchronous movement, unpredictable and frequent direction change, scattering and regrouping etc. In a swarm, every particle learns from his current experience and shared experience of other particles of the swarm. Kennedy and Eberhart [103] proposed PSO based on this hypothesis.

The social behaviour of particles synchronizes a collision-free movement in the search space towards the roost. For that, each particle matches its velocity with the nearest neighbors and maintains inter-individual distance in swarms. For maintaining the unanimous and unchanging direction random variable craziness is added in velocity. A roost is a location in the search space that attracts the particles until they reached there. It may be a food source, wire or a tree. For optimizing the movement towards the roost, each particle remembers their best value and share global best value with other particles of the swarm. Each particle synchronizes its flying movement based on the remembered value.

In computer simulation of PSO, all particles have its own velocity and position. Each particle remembers the personal best position (pbest) and shares the global best position (gbest) among all the particles for finding quality solutions. The particle's new position depends on the distance of its current position from pbest and gbest. The new velocity and new position of particles are calculated using equations 2 and 3. These equations have been developed after various development stages.

$$new\_velocity = current\_velocity + c_1 r_1 \times (pbest - current\_position) + c_2 r_2 \times (gbest - current\_position) \tag{2}$$

$$new\_positon = current\_position + new\_velocity \tag{3}$$

where $r_1$, $r_2$ are two random variables with range [0, 1] and $c_1$, $c_2$ are the learning factors.

The main control parameters in PSO are the number of particles, number of iterations and learning factors. Number of particles and number of iterations are the common control parameters. Learning factor and maximum velocity are algorithm-specific control parameters.

PSO is conceptually simple and computationally economical (in terms of speed and memory) and can be programmed with few lines code. It uses only few basic arithmetic operators. PSO does not use crossover or mutation operator like GA. The adjustment of pbest and gbest is conceptually similar to the crossover or mutation of GA.

There are more than two dozen PSO variants and its hybrid approach with other nature-inspired approaches like- GA, ABC has also been investigated [13, 227]. The different variants of PSO have different parameters like- inertia weight, learning factors, velocity clamping, acceleration constants and mutation operators [86]. The comparative study of various existing variants of PSO is shown in Table 4. In many problems, PSO gives better results than traditional optimization methods and even better than genetic algorithm. Some of the latest applications of PSO are:

software cost estimation [198], human motion tracking [186], data clustering [60], resource allocation in the cloud [135], online dictionary learning [217], capacitor placement problem in distribution system planning [115], vehicle routing problem [231], optimal power management based on driving condition for electric vehicles [32], robotic [137], inventory and location control in supply chain network [136], assembly line balancing [42] and brain MR image segmentation [113].

Table 4. Comparative study of PSO variants

| Reference | Algorithm | Algorithm/s compared with | Application |
|---|---|---|---|
| [104] | Binary PSO | - | De Jong's-1975 test bed |
| [63] | Regrouping PSO (RegPSO) | PSO | Ackley, Griewangk, Quadric, Rastrigin, Rosenbrock, Spherical and Weighted Sphere function |
| [240] | Cooperative Quantum-behaved PSO (CQPSO) | PSO and Quantum-behaved PSO (QPSO) | Distribution algorithms |
| [196] | Discrete PSO | Decimal Codification based GA (DCGA) | Transmission network expansion planning |
| [134] | Modified Binary PSO | PSO and BPSO | Cancer diagnoses |
| [199] | Personal Best Position PSO (PBPPSO) | PSO | 15 Scalable problems and 13 Non-scalable problems |
| [232] | Niche PSO | PSO | Target tracking |
| [233] | Neighborhood Search Barebones (NSBPSO) | Barebones PSO (BPSO) | Ship design |
| [93] | Immunity-Enhanced PSO (IEPSO) | DE and PSO | Structural damage detection |
| [220] | MOPSO | Well-known evolutionary multi-objective algorithms | Feature selection and classification |
| [109] | Guaranteed Convergence PSO (GCPSO) | Earlier reported approaches | Optimal power flows problem |
| [202] | Hybrid PSO-GA | GA and PSO | Closed-Loop Supply Chain (CLSC) network design |
| [218] | Hybrid GA-PSO | Basic GA and PSO | Welding robot path planning |

*4.2.3 Artificial Bee Colony (ABC)*
Intelligent foraging behaviour of honey bee inspired Karaboga [94] to propose a new swarm based algorithm. This is known as ABC. ABC obtains the swarm intelligent behaviour by self-organizing and division of labour concepts. Self-organizing includes- positive feedback, negative feedback, fluctuation and multiple interactions. Division of labour means each task is performed by specialized individuals. The swarm of honey bee has three basic components- food source, employee foragers and unemployed forgers. The honey bee searches food sources in intelligent and well-organized manner. When any bee finds a good food source based on nectar value, it shares collected information with other bees and the rest of the bees follow the same food source.

In ABC [95], the hive has three types of bees namely: employed bees, onlooker bees and scout bees. These bees have distinct role in hive. Scout bees search the food locations in the vicinity of the hive randomly. It becomes employed once it finds the location of a food source. Employed bees locate their food source; evaluate its amount of nectar and return to hive. It then starts dancing near the hive. This dance is known as waggle dance. The information exchange between the bees is an important process in ABC. The communication depends on the quality or richness of food source. The quality information of current sources is present in dance floor and it is measured by bee's dance duration. Based on the probability (or profitability) of food source, onlooker bee chooses a best food source.

During every search cycle, bees follow three basic steps [94, 97].

1. In first step, employed bees locate the food sources and measure the value of nectar.
2. In second step, employed bee shares the value of nectar with onlookers, and then a good food source is selected by onlookers.
3. In third step, scout bees explore the search area for new food sources.

In simulation of ABC, possible solution of problem is represented by food source. One employed bee is associated with every food source. The fitness of objective function is expressed by the amount of nectar or quality of a food source. The onlooker bees search the food source based on probability of source. The scouts explore the other food source without any guidance.

In ABC, employed bees find the new food source based on the neighboring food sources and the probability of new sources is calculated. Based on these probabilities, the onlooker bees compute the newly discovered food source position. The selection between new and current food source is based on greedy approach, i.e. one that gives the best solution. Finally, scout bees identify the abandoned food source and replace it by random food source. If successive iteration does not improve the probability of a food source, then it is an abandoned food source. The new solution position and probability are determined by equation 4 and equation 5 respectively.

$$new\_position = current\_position + rand \times (current\_position - current\_position\_k) \qquad (4)$$

where $rand$ is a random number within range [-1, 1] and random dimension index $k$ is selected between 1 to n.

$$Probability = \frac{fitness}{total\_fitness} \qquad (5)$$

The control parameters of ABC are the number of food source which same to the number of onlookers or employed bees, maximum number of cycles and the value of limit [97]. The number of food source and value of limit are algorithm-specific parameters. Limit is a control parameter to control the selection of food source, meaning that if the profitability of a food source is not improved within prefix trails then the food source is converted into an abandoned source. ABC is a simple, flexible and robust method. In ABC, exploration and exploitation process work in parallel. Onlooker and employed bees exploit the search space, while the scout bees explore the search space [94].

Many researchers have proposed various variants and modified ABC. A comparative study of various existing variants of ABC is shown in Table 5. Some of the recent applications of ABC are: travelling salesman problem [98], independent path and software test suite optimization [114], software effort estimation [70], economic dispatch problem [191], mobile robot path planning [34], load balancing in cloud [144], optimal placement problem in wireless sensor network [77], optimal power flow [2], image segmentation [29], energy aware routing in WSN [110], crack identification in beam [43] and FIR filter design [52].

Table 5. Comparative study of ABC variants

| Reference | Algorithm | Algorithm/s compared with | Application |
|---|---|---|---|
| [12] | Chaotic ABC (CABC) | ABC algorithm | Rosenbrock, Griewangk and Rastrigin function |
| [241] | Cooperative ABC (CABC) | ABC, PSO and Cooperative PSO (CPSO) | Sphere, Rosenbrock, Griewank, Rastrigin, Ackley and Schwefel function |
| [89] | Hybrid GA-ABC | GA, ABC and Conventional Gradient Descent Method | Proportional Integral (PI) speed controller of Permanent Magnet Synchronous Motor (PMSM) |
| [76] | Discrete ABC (DABC) | GA and ABC | Blocking Flow Shop (BFS) scheduling problem |
| [96] | Constrained ABC | GA, PSO, ABC, Homomorphous Mapping (HM) and Adaptive Segregational Constraint Handling EA (ASCHEA) | 13 linear, nonlinear and quadratic test functions |
| [53] | Hybrid ABC-SPSO | ABC and PSO | CEC05 benchmark functions |
| [9] | Multi-objective ABC (MOABC) | Multi-Objective EA based on Decomposition (MOEAD), Dynamical Multi-Objective EA (DMOEADD) and Multiple Trajectory Search (MTS) | Unconstrained and constrained test problems |
| [100] | Binary ABC (DisABC) | Binary DE (BinDE) and PSO | Uncapacitated Facility Location Problem (UFLP) |
| [18] | ABC with Levy Flight distribution (LFABC) | ACO, Standard ABC and Dynamic Harmony Search | Steel space frame design problem |
| [108] | Co-variance guided ABC (M-CABC) | MOABC | Portfolio optimization |

*4.2.4 Firefly Algorithm (FA)*

The intelligent flashing behaviour of fireflies has worked as a source of inspiration in development of new algorithms. Each firefly has a unique flashing pattern. It works as a signaling and communication mechanism between them and to attract the prey. It also acts as a protective warning system. Generally, the male and female fireflies attracted each other with a unique pattern of flashing for mating. The intensity of flashing light depends on the attractiveness of fireflies, the distance between the fireflies and the degree of absorption of the medium.

Yang [227] associates the flashing behaviour of fireflies with the objective function and proposes a new algorithm based on that which is known as FA. The FA has conceptual similarity with BFA (Bacterial Foraging Algorithms). In BFA, attraction between bacteria partly depends on fitness value and partly on the distance between them. But in FA, attraction is based on objective function and monotonic decay of attraction with distance. FA explores search space more efficiently than BFA.

The computer simulation of firefly algorithm depends on the following flashing rules of firefly [224, 225]:

1. The first rule states that, all fireflies are unisexual. For sexual activity, they are attracted to other fireflies.
2. According to the second rule, attractiveness is proportional to their brightness. Both being inversely proportional to their mutual distance between the fireflies. A less brighter firefly moves towards the more brighter firefly. If a brighter firefly is not found, then they will move randomly. Attraction also depends on the degree of absorption.
3. The third rule states that, the brightness of a firefly is equal to the fitness of objective function.

In FA, the attractiveness is calculated by brightness which is calculated with the association of the encoded objective function. The light intensity fluctuates due to distance and absorption by the media. The calculation of attractiveness and fluctuation in light intensity is very important. In general, the brightness (I) and attractiveness (β) of a firefly are calculated using equations 6 and 7 respectively. The movement of a firefly i towards the brighter firefly j is calculated using equation 8.

$$I(r) = I_0\, e^{-\gamma\, r^2}\ or\ \frac{I_0}{1 + \gamma r^2} \tag{6}$$

$$\beta(r) = \beta_0\, e^{-\gamma\, r^2}\, or\ \frac{\beta_0}{1 + \gamma r^2} \tag{7}$$

where $\gamma$ is the light absorption coefficient. $I_0$ and $\beta_0$ are the brightness and attractiveness at distance $r = 0$ respectively.

$$X_i = X_i + \beta_0 e^{-\gamma r_{ij}^2}(X_j - X_i) + \alpha \varepsilon_i \tag{8}$$

where the second and third term represents the attractiveness and randomization of $\alpha$ with randomization parameter $\epsilon_i$ respectively. For most implementations $\beta_o = 1$, $\alpha\ \epsilon\ [0,\ 1]$ and $\gamma\ \epsilon\ [0, \infty)$, but normally it lies between 0.01 and 100.

In FA, the control parameters are number of fireflies, number of iterations, light absorption coefficient and attractiveness. Light absorption coefficient and attractiveness are the algorithm-specific control parameters. Number of fireflies and number of iterations are the common control parameters. Due to variation in attractiveness, FA explored the search space efficiently. FA is capable of finding local and global optima simultaneously. Thus, FA is appropriate for parallel execution. FA exploits the search space better than GA and PSO [227].

A comparative study of various existing variants of FA is shown in Table 6. Some of the popular applications of FA are: load dispatch problem in economic emissions [14], travelling salesman problem [88], clustering algorithm [192], feature selection [21], image compression [81], image registration [238], manufacturing cell formation [190], image watermarking [128] and software effort estimation [71].

Table 6. Comparative study of FA variants

| Reference | Algorithm | Algorithm/s compared with | Application |
|---|---|---|---|
| [228] | Levy Flights FA (LFA) | GA and PSO | Michalewicz, Rosenbrock, De Jong, Schwefel, Ackley, Rastrigin, Easom, Griewank, Yang and Shubert function |
| [66] | Gaussian FA (GD- | PSO, FA and Time-varying | Sphere, Rosenbrock, Rastrigin, |

| | FF) | inertia weight PSO(PSO-TVIW) | Griewank and Ackley function |
|---|---|---|---|
| [209] | Elitist FA | FA | Seven benchmark problems |
| [30] | Binary Real Coded FA (BRCFF) | GA, PSO and Simulating Annealing (SA) | Unit Commitment Problem (UCP) |
| [85] | Parallel FA | GA, PSO and FA | Graphical Processing Unit (GPU) implementation |
| [6] | Hybrid Evolutionary Firefly Algorithm (HEFA) | GA, PSO, Evolutionary Programming (EP) and FA | Parameters optimization in a complex and nonlinear problem |
| [173] | Hybrid ACO–FA | Harmony Search (HS), PSO, GA, Particle Swarm Ant Colony Optimization (PSACO) | Several benchmark problems of unconstrained optimization |
| [67] | Chaos FA (CFA) | FA | Sphere and Griewank function |
| [229] | Multi-objective FA (MOFA) | Vector Evaluated GA (VEGA), NSGA-II, MODE and DEMO | Design optimization benchmarks in industrial engineering |
| [158] | Discrete Firefly Algorithm with the Standard GA | CPLEX and PSO | Capacitated Facility Location Problem (CFLP) |
| [99] | Hybrid Discrete Firefly Algorithm (HDFA) | PSO+SA, PSO+TS, MOPSO+LS and FL+EA | Flexible Job Shop Scheduling Problem (FJSP) |

## 4.3 Biological based Algorithms

The biological based algorithms are inspired by social behavioral pattern of biological sciences like botany and zoology. Botany includes the plant systems and zoology focuses in the living systems. Biological systems have many characteristics like robustness, adaptability and optimal decision. These characteristics work as motivational factor. The biologically inspired algorithms are described in detail next.

### 4.3.1 Shuffled Frog Leaping Algorithm (SFLA)

SFLA combines the principle of Shuffled Complex Evolution (SCE) and Particle Swarm Optimization (PSO) [62]. SFLA inherits the deterministic search approach from PSO and the random search approach from SCE. SFLA is inspired by the information sharing among the memetics of group.

Consider a swamp with frogs, in which stones are laid in different position on to which the frogs leap to find stone containing the maximum amount of food. During this process, frogs share the food information with others frogs, so that memes can be improved. The memes represent the traits of frog same as a genes in a GA [61]. The improvement of memes implies improvement in individual position of frogs. SFLA is based on this choreograph situation.

During the simulation of SFLA, the frogs are divided into several memeplexes (subgroups). Within the memeplexes, frogs share their experiences with the other members. This process is known as memetic evolution. This improves the quality of memes and improves the individual position of frogs towards the goal. Memetic evolution is repeated for a specific number of times. Later, frogs of memeplexes are reshuffled. This shuffling process improves the quality of memes by sharing the experience of frogs from different memeplexes.

SFLA was introduced by Eusuff and Lansey [61]. In SFLA, the population of virtual frogs is known as solutions and fitness value is known as performance. To begin with, all frogs are arranged according to their performance in decreasing order. After this, all frogs are partitioned into a number of memeplexes in such way that each memeplexes have an equal number of frogs.

Within each memeplex a local evolution is performed to find a local optimal solution. After a definite number of memetic evolution steps, all frogs are reshuffled for a global evolution to find a globally optimal solution. This process is repeated until stopping criteria not met like - number of fixed iterations or the desired solution meets. During the memetic evolution, the new position of the frogs is calculated by the equations 9 and 10.

$$change\_frog\_position = rand \times (frog_{best\_futness} - frog_{worst\_fitness})  \qquad (9)$$

$$new\_frog\_position = current\_frog\_position + Change\_frog\_position$$
$$where -D_{max} \leq change\_frog\_position \leq D_{max} \qquad (10)$$

where *rand* is a random number in the interval [0, 1] and the value of change frog position is within the range of the maximum permissible change in frog's position.

If new position of frog is better than previous frog position, then the worst frog replaced with new frog (solution). Otherwise, the same calculation is repeated with the global best frog (i.e. $frog_{best\_fitness}$ replace by $frog_{global\_best\_fitness}$). If there is no scope for improvement in the solution, then worst frog is replaced with a new randomly generated solution. The memetic evolution step is conceptually similar to PSO and the creation of memeplex and shuffling step is based on the SCE algorithm.

In SFLA, the control parameters are number of frogs and number of iterations, number of memeplexes, size of memeplex and number of evolution steps. The algorithm-specific control parameters are number of memeplexes, size of memeplex and number of evolution steps. The number of frogs and number of iterations are the common control parameters. SFLA is robust, fast for finding the solution and suitable for parallelization problem [62].

The comparative study of different existing variants of SFLA is shown in Table 7. The SFLA is applied in numerous applications. Some of the recent applications of SFLA include grid task scheduling [222], UAV flight controller problem [155], set covering problem [35], brain MR image segmentation [112], clustering in WSN [64], vehicle routing problem [122], resource scheduling in a cloud [126], manufacturing cell design problem [203], economic dispatch problem [185], job shop scheduling problem [117] and ground water calibration problem [61].

Table 7. Comparative study of SFLA variants

| Reference | Algorithm | Algorithm/s compared with | Application |
|---|---|---|---|
| [54] | SFLA with search acceleration factor (MSFL) | SFLA and GA | Construction project management (time-cost trade-off) |
| [153] | Modified SFL (MSFL) with adaptive coefficient | SFLA | Shubert, Hartmann-3, Shekel, Hartmann-6, Rosenbrock and Zakharov functions |
| [19] | Discrete SFL (DSFL) | SFLA, Discrete PSO (DPSO) and Binary GA (BGA) | Nonlinear and multimodal functions |
| [170] | Chaotic SFLA (CSFLA) | SFLA, Variants of GA and PSO | Sphere, Schwefel, Rosenbrock, Quadric, Rastrigin and Griewank function |
| [112] | MSFLA | 3D-Otsu thresholding with SFLA and GA | MR brain image segmentation |

*4.3.2 Flower Pollination Algorithm (FPA)*

Pollination is a natural process of biological evolution of plant. It is based on fertilization of seeds. Based on the pollination process, Yang [226] proposed a FPA. The main purpose of a flower in a plant is reproduction. The reproduction takes place via pollination process.

During pollination, the pollens are transferred from one flower to another flower with the help of pollinators like- birds, bees, insects, humans and others. Without pollinators pollination is not possible. Normally two forms of pollination take place- abiotic and biotic. During biotic pollination, pollens are transferred by pollinators like- birds, bees, insects and other animal. During abiotic pollination, there is no need of any pollinator. Wind and water helps in pollination. Pollination can also take place as self-pollination or cross-pollination. Self-pollination occurs in the same flower or different flower of the same plants while cross-pollination occurs between the flowers of different plant.

For simulation, FPA follows the basic considerations [226]:

1. Self-pollination and abiotic pollination comprise as local pollination.
2. Cross-pollination and biotic pollination comprise as global pollination.
3. Transition between the pollination is controlled by switching probability P.
4. The reproduction probability of flower is measured as flower constancy. It depends on the similarity of two flowers.

In local pollination, flower pollens move in very short range. This movement happens due to the physical closeness and other factors like wind, water, etc. In global pollination, flower pollens can move a long distance over a long range. It can be made by animals, birds or humans. The pollination process ensures the reproduction of the fittest. Normally local pollination has a higher fraction in overall pollination process. Flower constancy of pollination is calculated by equations 11 and 12 respectively.

$$new\_pollen = current\_pollen + L(current\_pollen - g_*) \qquad (11)$$

where $g_*$ is the current fittest solution and parameter $L$ is the strength of the pollination.

$$new\_pollen = current\_pollen + \epsilon(pollen_j - pollen_k) \qquad (12)$$

where $pollen_j$ and $pollen_k$ are pollens from the different flowers of the same plant (population). $\epsilon$ is a uniform distribution in [0, 1]. Pollen is a mimic of the flower constancy.

In FPA, the control parameters are number of pollens and number of iterations, switching probability and strength of the pollination. Switching probability and strength of the pollination are algorithm-specific control parameters. Number of pollens and number of iterations are the common control parameters. Most of the applications work well at switching probability 0.8.

The comparative study of various existing variants of FPA is shown in Table 8. Paper [20] and [33] present a review of the FPA and its applications. FPA has numerous applications such as image compression [101], economic load dispatch problem [154], retinal vessel segmentation in medical [57], life time of a node optimization in WSNs [194], feature selection [174], solving Sudoku puzzles [5], photovoltaic parameter selection in renewable energy [11], optimal placement in distributed system [172], economic and emission dispatch problem [3] and capacitor placement problem in electric power distribution system [4].

Table 8. Comparative study of FPA variants

| Reference | Algorithm | Algorithm/s compared with | Application |
|---|---|---|---|
| [230] | Multi-objective FPA (MOFPA) | Non-dominated Sorting GA (NSGA-II), Vector Evaluated GA (VEGA), Multi- | Design of a disc brake |

| | | objective DE (MODE), DE for Multi-objective optimization (DEMO), Strength Pareto EA (SPEA) and Multi-objective Bees algorithm (Bees) | |
|---|---|---|---|
| [5] | Hybrid FPA with Chaotic Harmony Search (FPCHS) | Harmony Search (HS) | Sudoku Puzzles |
| [90] | Hybrid FPA with K-Means (FPAKM) | K-Means and FPA | Data Clustering |
| [174] | Binary FPA | PSO, FA and Harmony Search (HS) | Feature selection |
| [138] | Modified FPA (MFPA) | FPA, Bat Algorithm, FA, GA and Simulated Annealing (SA) | 23 well-known optimization benchmark functions |

## 4.4 Science based Algorithms

Science based algorithms are based on a scientifically proven concept of physics or chemistry or mathematics. These concepts are the basic principles of the universe. The algorithms inspired by science are described next.

### 4.4.1 Gravitational Search Algorithm (GSA)

GSA is inspired by Newton's law of gravity and motion. Three types of mass explained in physics theory: active mass, passive mass and inertia mass. Based on this concept of mass, the law of gravity and law of motion are rewritten as- "The gravitational force ($F_{ij}$), acting on mass i by mass j, is proportional to the product of the active gravitational of mass j ($M_{aj}$) and passive gravitational of mass i ($M_{pi}$), and inversely proportional to the square distance (R) between them. The acceleration ($a_i$) is proportional to gravitational force ($F_{ij}$) and inversely proportional to the inertia mass of i ($M_{ii}$)."

Based on the above gravitational force theory, Rashedi et al. [168] proposed GSA. For simulation, each mass (agent) has four properties: position, inertial mass, active mass and passive mass. The position defines a solution of problem and masses are calculated by objective function. The heaviest mass is considered as an optimal solution [168]. Due to gravitational force, every agent attracts each other. Due to attraction all agents move towards the heavier agents (higher gravitational force). The movement produces acceleration in agents. The force between the agents depends on the product of their masses and distance between them instead of square of distance. Because, it gives better result than the square of distance. The force is calculated by equation 13. The acceleration is computed as proportion of total acting force on the agent and mass of the agent. The acceleration is calculated by equation 14. Due to acceleration, every agent has some velocity. Due to the velocity, the agents will update its position. The velocity and position of the agents are computed by equations 15 and 16.

$$force = gravitational\_const \times \frac{active\_mass \times passive\_mass}{distance} \tag{13}$$

$$acceleration = \frac{total\_acting\_force}{inertial\_mass} \tag{14}$$

$$new\_velocity = rand \times current\_velocity + acceleration \tag{15}$$

$$new\_position = current\_position + new\_velocity \tag{16}$$

where *rand* is a uniform random variable in the interval [0, 1].

GSA has improved performance capability in term of the exploration and exploitation [177]. In GSA, number of masses and position of the agent are common control parameters and the gravitational constant is an algorithm-specific control parameter.

Since the development of GSA, lots of modified variants have been proposed. These modifications enhance the incremental performance of GSA. The comparative study of different variants of GSA is shown in Table 9. In paper [177], author disused variants and application of GSA. The GSA has a number of good applications in various areas. Some of them are: IIR and rational filter modelling [169], parameter optimization of sensor monitoring to minimize the energy [175], routing and wavelength assignment problem in optical networks [176], PID controller problem [51], optimal power flow problem [50], economic and emission dispatch problem of power systems [195], finding the near-optimal base station in WSNs [157], energy efficient WSNs [148], optimal IIR filter designing [183], gas synthesis production problem [68], data clustering and classification [111] and heat and power economic dispatch problem [24].

Table 9. Comparative study of GSA variants

| Reference | Algorithm | Algorithm/s compared with | Application |
|---|---|---|---|
| [167] | Binary GSA (BGSA) | GA, Binary PSO (BPSO) | Seven unimodal, five multimodal, ten multimodal test functions with fix dimension |
| [78] | Multi-objective GSA with uniform mutation and elitist policy | MOPSO. MOGSA and Several Multi-objective EA (MOEAs) | Three multi-objective functions (MOP5, MOP6 and MOPCI) |
| [72] | Multi-Objective GSA (MO-GSA) | NSGA-II and Multi-objective GA for motif discovery (MOGAMOD) | Motif Discovery Problem (MDP) (DNA patterns) |
| [127] | Hybrid PSO and GSA (PSOGSA) | PSO and GSA | 23 benchmark functions |
| [87] | Hybrid GSA and Fuzzy Logic (GSA-FL) | Fuzzy logic control and Harmonic distortion | Controlling if active power filter |
| [199] | Hybrid genetic gravitational algorithm | GSA | Gait for the hexapod robot |
| [141] | Non-dominated Sorting GSA (NSGSA) | NSGA-II, MOGSA and MOPSO | Multi-objective benchmark problems (SCH, FON, POL, KUR, and ZDT) |
| [16] | Intelligent GSA based classifier (IGSA-classifier) | Swarm intelligence based and evolutionary classifiers | Pattern recognition problem and different benchmarks |
| [118] | Chaotic GSA (CGSA) | GA, PSO and GSA | Identifying the parameters of Lorenz chaotic system |
| [74] | Hybrid GSA and ABC (GSA-ABC) | ABC and GSA | Five benchmark functions |
| [214] | Gravitational Particle Swarm (GPS) | PSO and GSA | Seven unimodal, five multimodal, ten multimodal test functions with fix dimension |
| [234] | Niche GSA (NGSA) | State-of-the-art Niching algorithms | Unconstrained and |

| [91] | Hybrid PSO and GSA (HPSO-GSA) | PSO and GSA | constrained standard benchmark functions |
| | | | Economic Emission Load Dispatch (EELD) problems |
| [38] | Hybrid of Improved PSO and Improved GSA (Hybrid IPSO–IGSA) | IPSO and IGSA | Multi-robot path planning |

*4.4.2 Water Cycle Algorithm (WCA)*

The water cycle process and flow of streams in the real world are the source of inspiration for the WCA. This is a natural phenomenon. This process includes various activities like- evaporation, condensation, precipitation, transpiration, percolation and surface run-off [59].

In the real world, sources of water are rain, snow and groundwater. Open sources of water are created by rain. Stream and river originate in the mountain top where icy masses liquefy. The water absorbed by aquifer, is known as groundwater. Generally, streams and rivers move downhill on the surface. The streams combine into the river, rivers merge into the sea and finally, all streams and river become one into the sea. The water evaporates from open water bodies, like-rivers, ponds and lakes. The evaporation and transpiration of water produce clouds and water then comes back in the form of rain and snow. This observation inspired Eskandar et al. [59] to propose a meta-heuristic algorithm named the WCA.

In simulation of WCA [59], the population is known as raindrops and fitness value is known as cost of raindrop. The best raindrop is selected as a sea and some good raindrops are selected as a river and the remaining raindrops are selected as a stream. The number of rivers is a user parameter. The river /or sea absorbs the water from the stream based on the magnitude / intensity of the flow. The intensity is calculated by the equation 17. Intensity represents the number of streams which flow into a river or sea. Normally, the stream flows into river or directly into sea and river flows into the sea. The new position of a stream and river is calculated by equations 18 and 19. Same calculation is repeated with exchange the position of stream and river; river and sea. The best position in between both calculated position is selected as a next position of stream and river. Water evaporation is an important process in a water cycle algorithm which avoid from the rapid convergence. In evaporation condition, the distance between the sea and river controls the search intensity near the sea. If distance is lesser than a fixed value, then the evaporation process starts. If evaporation condition is satisfied, then raining starts. During raining process, new raindrops and streams originate at different places. The new position of these new streams is calculated by equation 20 which is similar as mutation operator of GA. Again, same process in repeated with new raindrops.

$$intensity\_flow = round\left\{\left|\frac{cost}{total\_cost}\right| \times no\_raindrop\right\} \qquad (17)$$

$$new\_pos\_stream = current\_pos\_stream + rand \times C \\ \times (current\_pos\_river - current\_pos\_stream) \qquad (18)$$

$$new\_pos\_river = current\_pos\_river + rand \times C \\ \times (current\_pos\_sea - current\_pos\_river) \qquad (19)$$

$$new\_stream = LB + rand \times (UB - LB) \qquad (20)$$

where $C$ is random variable between 1 and 2; $LB$ and $UB$ are the lower and upper bounds of problem's variable.

WCA always aims to find a globally optimal solution via effective exploration and exploitation [181]. WCA uses less number of insensitive user parameters, which means that WCA is capable of solving numerous optimization problems via fixed user defined parameters [181]. The number of rivers and evaporation condition are the algorithm-specific control parameters.

Over the last few years, various improved, modified and hybrid version of the WCA have been proposed by various authors. A comparative study of various existing variants of WCA is shown in Table 10. WCA has numerous applications in several varieties of optimization problems such as weight optimization problem of truss structure [58], optimal operation of reservoir system [75], water distribution system [182], power system stabilizer [69] and load frequency controller for power system [56].

Table 10. Comparative study of WCA variants

| Reference | Algorithm | Algorithm/s compared with | Application |
|---|---|---|---|
| [59] | WCA | GA, DE, HS, Hybrid PSO and TLBO | Three-bar truss problem, Speed reducer problem, Pressure vessel design problem, Tension/ compression spring design problem, Welded beam design problem, Rolling element bearing design problem, Multiple disk clutch brake design problem |
| [180] | Multi-objective WCA (MOWCA) | NSGA-II, MOPSO, Micro-GA, Elitist-mutation Multi-objective PSO (EM-MOPSO), Hybrid Quantum Immune Algorithm (HQIA) | Four-bar truss design problem, Speed reducer problem, Disk brake design problem, Welded beam design problem, Spring design problem, Gear train design problem |
| [179] | WCA with Evaporation Rate (ER-WCA) | WCA, PSO, DE and BFO | Constrained and unconstrained optimization problems |
| [79] | Chaotic WCA | WCA, PSO and Variants of PSO | Several benchmark problems and training of NNs |

## 4.5 Other Algorithms

Sometimes it is hard to categorize the some algorithms in the above discussed classification; because they are inspired by any other natural phenomena like- teaching methodology, winning tendency, etc. In such case, these algorithms are put under in this category. Next section describes the teaching learning based optimization and Jaya algorithm.

### 4.5.1 Teaching Learning Based Optimization (TLBO)

The concept of TLBO is based on the teaching-learning methodology of class of learners. In this, various types of learning are possible like- learning from teachers, self-learning of learners, group learning of learners, learning from other learner which has higher knowledge, learning from assignments and examination, etc. Rao [166] considers the learning from teachers and learning from other learner for the teacher learning based optimization. During teacher's learning, the teacher teaches the learners and learner increases his or her knowledge from teacher. In learning from other learner, learners exchange their knowledge with other classmate which has higher knowledge, with the aim to increase his or her knowledge. The aim of this teaching-learning methodology is that learners should increase their knowledge and score higher grades.

For computer simulation, the TLBO process is performed in two phases [160, 164]: "Teacher phase" and "Learner phase". In teacher phase, the best learner of the class is selected as a teacher. The teacher emphasizes to increase the mean result of the class. In the learner phase, every learner increases his knowledge by interacting with other learner of class. This interaction among the learners happens randomly for enhancing knowledge of learners.

In TLBO, class of learners is considered as a population, number of subjects is considered as a design variable and a result is considered as a fitness value. In the teacher phase, first the mean result of the class is computed and then the difference mean of the class is calculated by equation 21. The difference mean is the difference between the best learner's results and mean of class result. The new solution depends on the current solution and the difference mean of the class. The new solution is calculated using equation 22. During the learner phase, two learners are selected randomly and then the new solution is calculated using equations 23 and 24.

$$difference\_mean = r \times (best\_learner - T_F \times mean\_result) \tag{21}$$

where $T_F$ is the teaching factor either 1 or 2 and $r$ is the random number in the range [0, 1].

$$new\_sol = current\_sol + difference\_mean \tag{22}$$

$$new\_sol = current\_sol + rand \times (current\_sol\_P - current\_sol\_Q), if\ P > Q \tag{23}$$

$$new\_sol = current\_sol + rand \times (current\_sol\_Q - current\_sol\_P), if\ P < Q \tag{24}$$

where $P$ and $Q$ are the randomly selected learners such that $P \neq Q$.

At the end of both phases, the new solution is selected only when it gives better result. TLBO does not require any dependent parameters. It requires only regular parameters like- number of learners, number of subjects and number of iterations.

The comparative study of the various existing variants of TLBO is shown in Table 11. TLBO has been successfully functioning in various areas of engineering and science. A large number of applications were presented by various researchers in the recent past. Some of the well known applications of TLBO are: data clustering problem [189], electric power dispatch problem [123], machine process parameter optimization [113], IIR filter designing [200], optimal capacitor placement problem [204], flow shop and job shop scheduling problem [23], PID-controller [184], software effort estimation [105] and optimizing the order neural network [139].

Table 11. Comparative study of TLBO variants

| Reference | Algorithm | Algorithm/s compared with | Application |
|---|---|---|---|
| [28] | Cooperative Co-evolutionary TLBO (CC-TLBO) | Modified TLBO (m-TLBO) | High dimensional problems |
| [165] | Elitist TLBO | TLBO, DE and EP | Constrained benchmark functions and optimization problems of the industrial environment |
| [73] | Multi-objective TLBO (MO-TLBO) | Multi-objective EA | Motif Discovery Problem (MDP) in Bioinformatics |
| [125] | Multi-objective TLBO based on Decomposition | Multi-objective EA based on decomposition (MOEA/D) | Reactive power handling problem |

|        |                                                      | (MOTLA/D)                                                                                                                                                               |                                                                                    |
|--------|------------------------------------------------------|-----------------------------------------------------------------------------------------------------------------------------------------------------------------------|------------------------------------------------------------------------------------|
| [92]   | Hybrid model of DE and TLBO (hDE-TLBO)               | Non-dominated scheduling schemes for the MOSOHTS                                                                                                                       | Optimal hydro-thermal scheduling (MOSOHTS)                                          |
| [215]  | Harmony Search Based Teaching Learning (HSTL)        | Harmony Search (HS) and complex benchmark functions                                                                                                                   | Unconstrained Optimization Problems                                                 |
| [171]  | Binary TLBO (BTLBO)                                   | PMU placement methods                                                                                                                                                  | Optimal placement of Phasor Measurement Units (PMU) placement problem              |
| [10]   | Hybrid of TLBO and Harmony search                    | Harmony Search (HS)                                                                                                                                                    | Designing of steel space frames                                                     |
| [119]  | Discrete TBLO (DTLBO)                                 | Different versions of DTLBO                                                                                                                                             | Flow shop rescheduling                                                              |
| [82]   | Teaching-Learning based Cuckoo Search (TLCS)         | Well-known constrained engineering design problems                                                                                                                    | Constrained optimization problems                                                   |
| [83]   | TLCS with Lévy flight                                 | TLCS                                                                                                                                                                   | Structure designing machine                                                         |
| [216]  | Hybridized TLBO with DE (TLBO–DE)                    | BAT, CS, Artificial Cooperative Search (ACS), Backtracking Search (BS), Melody Search (MS), Quantum behaved PSO (QPSO), and Intelligent Tuned HS (ITHS)               | Proton exchange membrane fuel cell (PEMFC) problem                                  |
| [150]  | Multi-objective Improved TLBO (MO-ITLBO)             | MO-TLBO                                                                                                                                                                | CEC 2009 standard test problems                                                     |

After the wide acceptance and popularity of TLBO, Rao [159] proposed a comparatively simpler algorithm than the TLBO, which has lesser computational steps and equally powerful as other algorithms.

### 4.5.2 Jaya Algorithm

Jaya algorithm [159] is based on the concept in which a solution moves towards the best solution and discards the worst solution. It always tries to find an optimal solution therefore victorious. That's why it is called Jaya algorithm because of the word 'JAYA' is derived from the word 'victory'. It is simple yet powerful algorithm.

During the simulation of Jaya, first the best and worst solutions are computed, then the current solution is modified by the equation 25 and the new solution is produced.

$$new\_sol = current\_sol + r1(best\_sol - |current\_sol|) - r2(worst\_sol - |current\_sol|) \qquad (25)$$

where $r1$ and $r2$ are the random number in the range [0,1].

Same as TLBO, Jaya does not have any dependent parameters. It has only regular parameters like- number of candidates (population size), number of design variable and number of iterations.

Jaya is a very recently introduced algorithm, so only limited variants have been found to date. The comparative study of existing variants of Jaya is shown in Table 12. The Jaya has been fruitfully applied in various areas of engineering and science like- micro-channel heat sink problem [161], surface grinding process optimization problem [163], nano-finishing process optimization [162], Tea Category Identification problem (TCI) [239], load dispatch problem [25], distributed energy resource distribution problem [213], Carbon Fibre-Reinforced Polymer (CFRP) [7], production scheduling problems [156] and photovoltaic parameter selection [132].

Table 12. Comparative study of Jaya variants

| Reference | Algorithm | Algorithm/s compared with | Application |
|---|---|---|---|
| [159] | Jaya Algorithm | Homomorphous Mapping (HM), Simple Multi-member Evolution Strategy (SMES), GA, DE, ABC, PSO, TLBO, Biogeography based Optimization (BBO) and Heat Transfer Search (HTS) | Constrained and unconstrained functions |
| [239] | Jaya algorithm and Fractional Fourier entropy | Various neural network based methods | Tea-Category Identification (TCI) problem |
| [156] | Multi-objective Jaya | GA and DE | Master Production Scheduling (MPS) |

## 5 COMPARATIVE STUDY OF NATURE-INSPIRED ALGORITHMS

Table 13 illustrates a tabular study of all discussed nature-inspired algorithms in terms of source of inspiration, objective function or basic operators, common control parameters, algorithm-specific parameters and main features. All these algorithms are arranged in chronological order.

Table 13: Summary of various nature-inspired algorithms

| S. No. | Algorithm and Develop by | Source of inspiration | Objective function / Basic operators | Common Control Parameters | Algorithm-Specific Control Parameters | Features |
|---|---|---|---|---|---|---|
| 1 | Genetic Algorithms (GA) – 1975 John H. Holland et al. | Darwin's theory of biological evolution | Crossover, Mutation and Selection | Population size, Number of generation | Crossover probability, Mutation probability, Chromosome length and encoding technique | Ability of explore and exploit simultaneously, Powerful, Robust |
| 2 | Ant Colony Optimization (ACO) – 1992 Macro Dorigo | Cooperative behaviour of real ants | Pheromone amount, Trail evaporation | Number of ants, Number of iterations | Pheromone evaporation rate, Heuristic information, Amount of reinforcement | Finding good paths through graph or tree |
| 3 | Particle Swarm Optimization (PSO) – 1995 | Social behaviour of creatures like-bird flocking or | Velocity and Position of particles new_velocity = current_velocity + $c_1 \times r_1 \times$ (pbest – | Number of particles, Number of iteration | Inertia weight, Learning Factors, Maximum velocity | Very simple concept, Computationally inexpensive |

| | | | | | |
|---|---|---|---|---|---|
| | Dr. James Kennedy and Dr. Russell Eberhart | fish schooling | current_position) + $c_2 \times r_2 \times$ (gbest − current_position) new_position = current_position + new_velocity | | | |
| 4 | Shuffled Frog Leaping Algorithm (SFLA) − 2003 Muzaffar Eusuff and Kevin Lansey | Leaping and shuffling behaviour of frogs | Replacement, Shuffling and Frog's position change_frog_position = rand× ($\text{frog}_{best\_fitness}$ − $\text{frog}_{worst\_fitness}$) new_frog_position = old_frog_position + change_frog_position | Number of frogs, Number of iterations | Number of memeplexes, Size of memeplexes, Number of evolutionary steps | Simple method with less computation, Stable convergence and high quality solution |
| 5 | Artificial Bee Colony (ABC) − 2005 Dr. Dervis Karaboga | Foraging behaviour of honey bee | Reproduction, Replacement of bee, Selection new_position = current_position + $\phi$ × (current_position − current_position_k) | Maximum cycle Number (MCN) | Number of food source, Value of limit | Relatively fast, Robust search process, Simple and flexible |
| 6 | Firefly Algorithms (FA) − 2007 Xin-She Yang | Intelligent flashing behaviour of fireflies | Brightness (light intensity), Attractiveness and Movement of firefly new_ff_i = old_ff_i + βο $e^{-\gamma r2ij}$ (old_ff_j − old_ff_i) + $\alpha\epsilon_i$ | Number of fireflies, Number of iterations | Attractiveness, Light absorption coefficient | High convergences rate, robust, Finds good optimum solutions in less number of iterations |
| 7 | Gravitation al Search Algorithm (GSA) − 2009 Esmat Rashedi et al. | Newton's law of gravity | Mass, Force, Acceleration, Velocity and agent's position new_velocity = rand × current_velocity + acceleration new_position = old_position + new_velocity | Number of masses, Position of agents, Number of iteration | Gravitational constant | High performance |
| 8 | Environme ntal Adaption Method (EAM) − 2011 Dr. K. K. | Improved version of Darwin principle | Adaption, Mutation and Selection new_sol = (α × (current_sol) ^(fitness(old_sol)/avg _fitness) + β) / $2^L$-1 | Population size, Number of generation | Mutation probability, Number of bits for individual | Adapt environment al condition |

| | | | | | |
|---|---|---|---|---|---|
| | Mishra | | | | |
| 9 | Teaching Learning Based Optimizati on (TLBO)- 2011 R. Venkata Rao et al. | Teaching-learning methodolog y | Teacher phase: difference_mean = r×(best_learner − $T_F$×mean_result) new_sol = current_sol + difference_mean Learner phase: new_sol = current_sol + r×(current_sol_P − current_sol_Q) | Population size, Number of iterations, Teaching Factor | NA | No algorithm-specific parameter required, efficiency in term of less number of function evaluations |
| 10 | Flower Pollination Algorithms (FPA) – 2012 Xin-She Yang | Flower pollination process of flowering plants | Global pollination: new_pollen = old_pollen + L× (old_pollen − current_best_pollen) Local pollination: new_pollen = old_pollen + ε × (old_random1_pollen − old_random2_pollen) | Number of pollens, Number of iterations | Switch probability, Strength of pollination | Simple, flexible, efficient, Higher convergence rate |
| 11 | Water Cycle Algorithm (WCA) – 2012 Hadi Eskandar et al. | Water cycle process and movement of water streams | new_pos_stream = current_pos_stream + rand×C× (current_pos_river − current_pos_stream) new_pos_river = current_pos_river + rand×C× (current_pos_sea − current_pos_river) | Number of raindrops, Number of iteration | Number of river, Evaporation condition, Number of design variable | Not trapped in local solutions |
| 12 | Jaya Algorithm – 2016 R. Venkata Rao | Victory (Win) | Move in the direction of best solution and dispose of the worst solution new_sol = current_sol + $r_1$×(best_sol - \|current_sol\|) – $r_2$× (worst_sol - \|current_sol\|) | Population size, Number of iterations | NA | Simple, No algorithm-specific parameter required, Yet powerful |

## 6   CONCLUSION

Nature-inspired algorithms are the recent evolution from GA. These are highly efficient algorithms and produce a near optimal solution for real-world optimization problems. The monumental impact of these algorithms is attributed to their widespread use for solving a vast variety of problems. We have presented a systematic review of various nature-inspired algorithms. Given that no algorithm proves their excellency in solving all the optimization problems. It may provide superior performance for some problems while it may perform poorly for other problems.

Many times, the characteristic of the problem may affect the performance of algorithms. Among the multitude of known optimization techniques, GA and PSO are most widely used. PSO is much simpler than the GA because it does not use crossover/mutation operators. ACO is efficient for the graph-based and tree-based optimization problem. EAM is the technique that adapts to current environmental changes, thus giving better performance than other techniques for solving the constrained and unconstrained optimization problems. Both FA and FPA show higher convergence rate and they improve the timing and result performance in various fields. Less number of algorithm-specific parameters is the characterizing feature of TLBO. With this technique, an optimal solution can be obtained in a comparatively lesser number of iterations. TLBO requires less computational effort for large scale problems. WCA offers efficient solutions than others in terms of the computational cost. Jaya is a straightforward and equally capable technique as other techniques. Also, it is independent of algorithm-specific parameters. The algorithms that have been recently proposed are yet to be explored in their application area. This comprehensive review of all known optimization algorithms can be used as a source of information for further research. Our aim is to encourage and pave way for upcoming researchers in assisting them to identify or develop novel and efficient optimization algorithms for tackling large scale real-world problems.

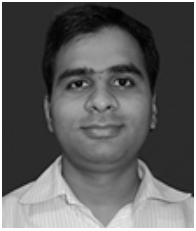

**Rohit Kumar Sachan** received B.Tech in Computer Science and Engineering (2008) from Galgotias College of Engineering and Technology, Greater Noida, India and M.Tech in Computer Science and Engineering (2012) from Amity University, Noida, India. He received Ph.D degree (2021) from Motilal Nehru National Institute of Technology Allahabad, Allahabad, India. Currently he is working as Sr. Project Engineer in C3i Center, Department of Computer Science and Engineering, Indian Institute of Technology Kanpur, Kanpur, India.

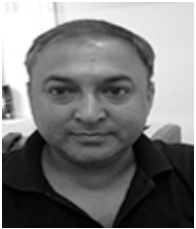

**Dharmender Singh Kushwaha** received B.E (Bachelor in Engineering) degree in Computer Science and Engineering from University of Pune, Maharasta, India, in 1990. He was awarded Gold Medal in M.Tech. (Computer Science and Engineering) from Motilal Nehru National Institute of Technology Allahabad, Allahabad, India. He received Ph.D degree from Motilal Nehru National Institute of Technology Allahabad, Allahabad, India. Currently he is working as Professor in Department of Computer Science and Engineering, Motilal Nehru National Institute of Technology Allahabad, Allahabad, India.